\documentclass[11pt, letterpaper, colorlinks=true, allcolors=blue]{article}
\usepackage[utf8]{inputenc}
\usepackage[T1]{fontenc}

\usepackage[margin=1in]{geometry}
\usepackage{setspace}

\usepackage[affil-it]{authblk} 
\usepackage{orcidlink} 

\usepackage{amsmath, amssymb, bm}
\usepackage{textcomp} 

\usepackage{graphicx}
\usepackage{booktabs}
\usepackage{caption}
\usepackage{subcaption} 
\usepackage{float}  
\usepackage{placeins}

\usepackage{siunitx}

\usepackage[numbers,sort&compress]{natbib}

\usepackage{lineno}
\usepackage{todonotes}

\usepackage{url}
\usepackage{hyperref}
\usepackage{xr}

\usepackage{siunitx}

\usepackage{xcolor}

\definecolor{revisioncolor}{named}{blue} 
\definecolor{revtwocolor}{HTML}{8B008B}
\definecolor{revchristelcolor}{HTML}{008080}

\newcommand{\revtwo}[1]{\textcolor{black}{#1}}

\newcommand{\rev}[1]{\textcolor{black}{#1}}
\newcommand{\revr}[1]{\textcolor{black}{#1}}
\newcommand{\revchristel}[1]{\textcolor{black}{#1}}

\title{AIFL: A Global Daily Streamflow Forecasting Model Using a Deterministic LSTM Pre-trained on ERA5-Land and Fine-tuned on IFS}

\author[1]{Maria Luisa Taccari}
\author[1]{Kenza Tazi}
\author[2]{Oisín M. Morrison}
\author[2]{Andreas Grafberger}
\author[1]{Juan Colonese}
\author[1]{Corentin Carton de Wiart}
\author[1]{Christel Prudhomme}
\author[1]{Cinzia Mazzetti}
\author[1]{Matthew Chantry}
\author[1]{Florian Pappenberger}

\affil[1]{European Centre for Medium-Range Weather Forecasts (ECMWF), Reading, United Kingdom}
\affil[2]{European Centre for Medium-Range Weather Forecasts (ECMWF), Bonn, Germany}

\date{\today}

\begin{document}
\maketitle


\begin{abstract}
Reliable global streamflow forecasting is essential for flood preparedness and water resource management, yet data-driven models often suffer from a performance gap when transitioning from historical reanalysis to operational forecast products. This paper introduces AIFL (Artificial Intelligence for Floods), a deterministic LSTM-based model designed for global daily streamflow forecasting. Trained on 18,588 basins curated from the \revr{Caravan} dataset, AIFL utilises a two-stage transfer-learning strategy to bridge the reanalysis-to-forecast domain shift. The model is first pre-trained on 40 years of ERA5-Land reanalysis (1980--2019) to capture robust hydrological processes, then fine-tuned on operational Integrated Forecasting System (IFS) forecasts (2016--2019) to adapt to the specific error structures and biases of operational numerical weather prediction. \rev{Ablation experiments confirm that this two-stage approach outperforms both a naive IFS-only baseline and a mixed-forcing single-stage alternative.} To our knowledge, this is the first global model trained end-to-end within the \revr{Caravan} ecosystem. On an independent temporal test set (2021--2024), AIFL achieves high predictive skill with a median modified Kling--Gupta Efficiency ($\text{KGE}'$) of 0.66 and a median Nash--Sutcliffe Efficiency (NSE) of 0.53. Benchmarking results show that AIFL achieves comparable accuracy to current state-of-the-art global systems. The model \rev{provides a streamlined and operationally robust baseline} for the global hydrological community.
\end{abstract}


\section{Introduction}

Global-scale streamflow forecasting is a critical capability for disaster risk reduction, supporting humanitarian aid, water resource management, and climate adaptation. The European Centre for Medium-Range Weather Forecasts (ECMWF) has long served as a central hub for these efforts, providing the computational backbone for the Copernicus Emergency Management Service’s Global Flood Awareness System (GloFAS) \cite{alfieri2013glofas}. Historically, hydrological forecasting has relied on a spectrum of approaches ranging from empirical relationships to complex, physically based frameworks \cite{singh2018hydrologic}. GloFAS traditionally relies on coupling numerical weather predictions (NWP) with process-based hydrological models to generate operational alerts. These models require rigorous calibration to link parameters to global geophysical maps—such as land cover, topography, and soil texture—using in-situ river discharge observations and forcing data like ERA5 \cite{prudhomme2024}. In GloFAS, this process is further enhanced by regionalization methods that transfer parameters from gauged "donor" catchments to ungauged regions based on geographical and climatic similarity \cite{grimaldi2023}. However, while these process-based models \revr{can provide valuable estimates where direct observations are unavailable}, they face significant challenges in representing complex hydrological processes accurately and are often limited by the quality and spatial resolution of climate-weather forcing variables \cite{prudhomme2024}. Although data-driven models share \rev{some of} these forcing-data limitations, the high computational demand of global-scale process-based simulations has motivated increasing interest in \revr{machine learning (ML)} alternatives that can deliver fast, scalable inference while maintaining competitive predictive skill.

The application of \revr{ML} to Earth system forecasting has recently accelerated, first transforming meteorology. ECMWF is pioneering this shift with the Artificial Intelligence Forecasting System (AIFS), a graph neural network-based model that now competes with physics-based NWP in medium-range accuracy \cite{lang2024aifs}. As detailed by Moldovan et al. \cite{moldovan2025update}, AIFS has transitioned to fully operational status and is expanding its capabilities beyond atmospheric variables to include land-surface outputs such as runoff, signalling a convergence of meteorological and hydrological ML capabilities.

A similar paradigm shift has occurred in hydrology \cite{slater2025challenges}. Kratzert et al. \cite{kratzert2024never} argued that ML models typically outperform traditional approaches for river discharge prediction when trained on large, diverse datasets rather than single basins. This hypothesis has been validated by the widespread adoption of Long Short-Term Memory (LSTM) networks, which have demonstrated the ability to learn universal hydrological behaviours, outperforming regionally calibrated process-based models and enabling accurate prediction in ungauged basins \cite{kratzert2019nse}. This success has spurred a diverse family of advanced architectures, such as Hydra-LSTM \cite{ruparell2025hydra}, which employs a semi-shared architecture to improve multi-basin prediction, and MC-LSTM \cite{wang2025investigating}, which integrates mass conservation constraints directly into the network structure.

Most state-of-the-art approaches employ a lumped formulation \cite{nearing2024global}, where the LSTM operates on inputs spatially aggregated over the entire catchment, including both time-varying meteorological forcings and static attributes such as topography, soil properties, and land cover. By collapsing the spatial distribution of these features into basin-wide aggregates, these models inherently overlook sub-catchment heterogeneity and the internal dynamics of river routing and lateral soil water redistribution. This structural simplification limits the utility of such models for tasks requiring fine-scale flow propagation or the representation of discrete hydrological features such as lakes, dams, and barrages \cite{scholz2025}.

To bridge this gap, a new generation of spatially explicit architectures has emerged. These range from implicit routing methods that learn flow propagation end-to-end \cite{moshe2020, shamseddin2025rivermamba} to hybrid frameworks that combine ML-based runoff generation with physically inspired transport schemes \cite{yang2025, bindas2024, kraft2025}. For instance, the DROP (Deep Runoff Prediction and propagation) framework \cite{kraft2025_drop} addresses these structural limitations by coupling drainage-unit scale LSTMs with a routing module, allowing for a more transparent and physically interpretable modulation of flow. However, the path to global implementation remains uneven; while some hybrid methods offer high fidelity, they often entail substantial computational overhead or rely on detailed river connectivity data at a level of detail that is not yet globally available. Consequently, deploying these spatially resolved methods in real-time operational systems remains a significant technical challenge

New global forecasting capabilities have emerged from these data-driven advances. Google's global flood forecasting model established a \revr{strong data-driven} benchmark for reliability in ungauged watersheds \cite{nearing2024global} using a specialized hindcast-forecast architecture. However, a critical operational gap remains for the broader research community: the mismatch between training data (historical reanalysis) and inference data (operational forecasts). Standardised datasets like CAMELS \cite{newman2015development} and \revr{Caravan} \cite{kratzert2023caravan} predominantly rely on meteorological reanalysis data (e.g., ERA5). Operational forecasting requires driving models with NWP forecasts, such as the ECMWF Integrated Forecasting System (IFS), which exhibit different error structures. This distribution shift often leads to degraded operational performance when models trained on "perfect" reanalysis are exposed to real-time forecast noise \cite{wang2025hydrodiffusion}. Addressing this challenge, Konold et al. \cite{konold2025biascast} demonstrated that a "domain shift" occurs when transitioning from reanalysis to forecast products, resulting in a significant reduction in predictive skill if not explicitly mitigated.

Importantly, this limitation is \rev{largely} orthogonal to architectural complexity. Even state-of-the-art models—whether spatially aggregated (lumped), connectivity-aware (graph-based), or those integrating physical constraints and process-based structures (hybrid)—are fundamentally constrained by the characteristics of their forcing data. Addressing this reanalysis-to-forecast domain shift is therefore a prerequisite for reliable operational deployment. Leveraging the recently introduced \revr{Caravan} MultiMet dataset \cite{shalev2025caravan}, we introduce AIFL (Artificial Intelligence for Floods). Unlike previous works that utilise complex probabilistic or explicit graph-based connectivity, AIFL utilises a standard, deterministic LSTM architecture trained on the entire \revr{Caravan} dataset (over 18,000 basins) to provide a scalable baseline.

The primary contribution of AIFL is a novel two-stage training strategy designed to solve the reanalysis-to-forecast domain shift. Inspired by findings that fine-tuning pre-trained models improves generalisation \cite{ryd2025fine}, we apply this concept to the temporal and data-source domain. We first pretrain the model on 40 years of ERA5-Land reanalysis to learn robust physical processes, and then fine-tune it on IFS control forecasts to adapt to operational biases. This approach offers \rev{an independent} baseline for global operational flood forecasting, with AIFL designed from the outset for integration into real-time forecasting workflows at ECMWF.

The remainder of this paper is organized as follows: Section~\ref{sec:data} details the data curation process; Section~\ref{sec:methodology} describes the model architecture and the two-stage training strategy; Section~\ref{sec:results} evaluates the model's predictive skill; and Section~\ref{sec:conclusion} concludes with future directions.

\section{Data Curation \& Experimental Design}
\label{sec:data}

\begin{figure}[htbp] \centering \includegraphics[width=0.7\textwidth]{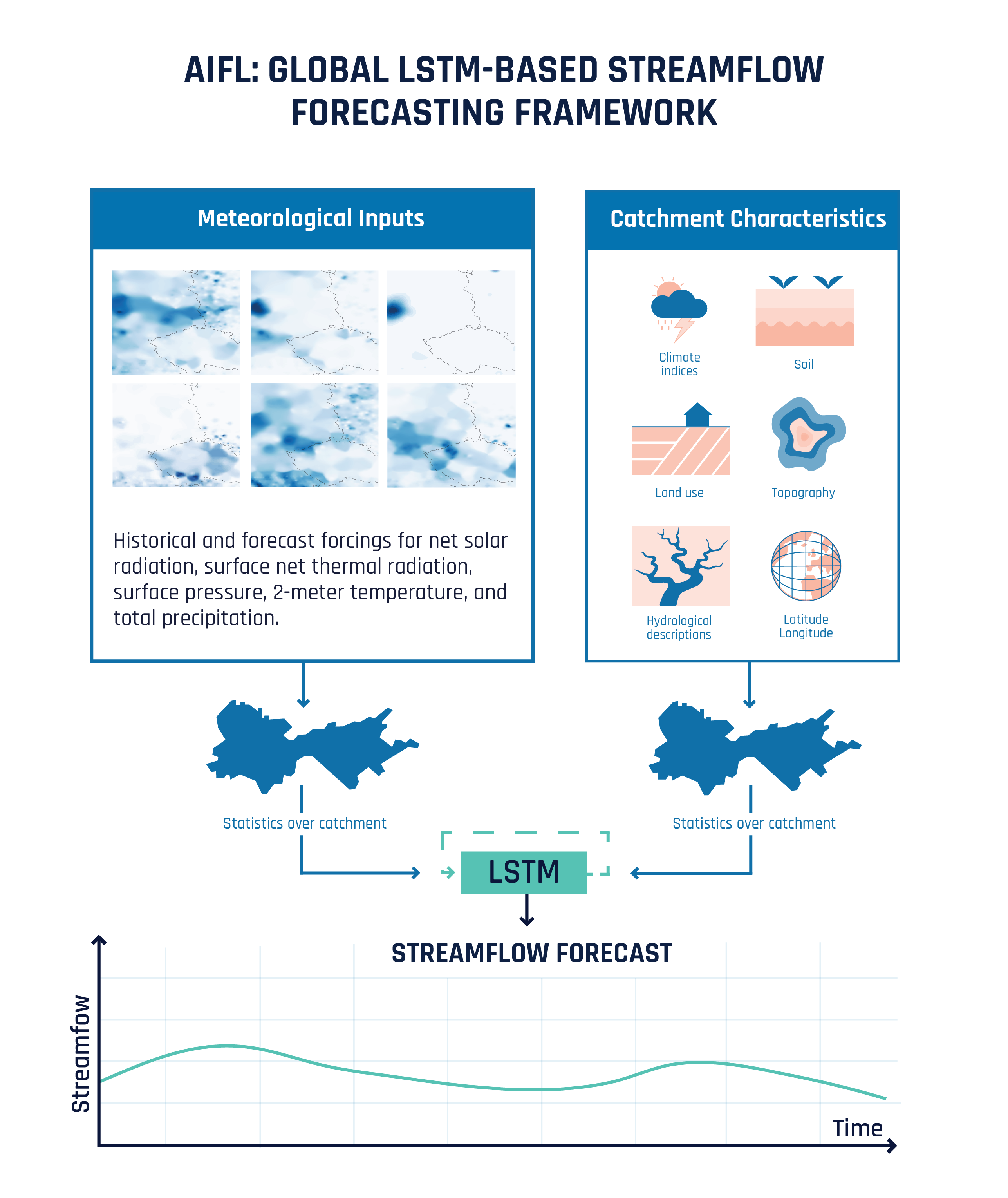} \caption{Schematic of the AIFL framework. The model architecture uses separate Multi-Layer Perceptron (MLP) embedding layers for static and dynamic inputs, feeding a shared LSTM core that processes a 170-day hindcast-window to generate 10-day forecasts. The training strategy transitions from ERA5-Land reanalysis pre-training to IFS forecast fine-tuning to resolve domain shifts.} \label{fig:model_diagram} \end{figure}

This section describes the data sources and processing steps, used to develop the AIFL model. The focus is on constructing a globally consistent, non-redundant training set that supports a two-stage reanalysis-to-forecast learning strategy. The AIFL framework (shown in Figure~\ref{fig:model_diagram} and detailed further in Section~\ref{sec:methodology}) utilises separate feedforward embedding networks to process static landscape attributes and dynamic meteorological forcings. These inputs are integrated by an LSTM core to produce streamflow forecasts; this architecture necessitates the specific data curation steps outlined below.

\subsection{Datasets and Target Variable}
The model is trained to predict daily streamflow in the form of specific discharge (\unit{mm.d^{-1}}), utilizing station-based observations from the \revr{Caravan} dataset v1.5 \cite{kratzert2023caravan} and its extensions, including CAMELS-US \cite{addor2017camels}, CAMELS-AUS \cite{fowler2021camelsaus}, CAMELS-BR \cite{chagas2020camelsbr}, CAMELS-CH \cite{hoge2023camelsch}, CAMELS-CL \cite{alvarez2018camelscl}, CAMELS-DE \cite{loritz2024camelsde}, CAMELS-DK \cite{koch2022camelsDK}, CAMELS-ES \cite{casado2023camelses}, CAMELS-GB \cite{coxon2020camelsgb}, HYSETS \cite{arsenault2020hysets}, LamaH-Ice \cite{helgason2024lamahice}, Caravan-Israel \cite{morin2024caravanisrael}, and GRDC-Caravan \cite{farber2025grdc}. Meteorological forcing data are provided by the Caravan-MultiMet extension \cite{shalev2025caravan}.
Specific discharge—defined as discharge normalised by the catchment drainage area—simplifies the learning task by aligning the units of the target variable with meteorological inputs such as precipitation. This design choice avoids the network learning linear basin-area scaling through complex transformations, while still incorporating basin size as an explicit input through the static attributes to capture mechanisms such as basin storage capacity, lag time, or concentration time.

\subsection{Deduplication and Quality Control}
The raw \revr{Caravan} dataset includes 22,371 river gauges. Because \revr{Caravan} aggregates data from multiple independent providers---including the Hydrometeorological Sandbox of \'{E}cole de \revr{T}echnologie \revr{S}up\'{e}rieure (HYSETS) \cite{arsenault2020hysets}, the Catchment Attributes and Meteorology for Large-sample Studies (CAMELS) dataset \cite{addor2017camels}, and the GRDC-Caravan extension \cite{farber2025grdc} which integrates data from the Global Runoff Data Centre (GRDC)---it lacks a unified quality-control standard and inherently contains spatial overlaps among its constituent datasets. While closely spaced river gauges along a stream are standard practice for flow monitoring, retaining gauges with near-identical drainage areas without filtering leads to information leakage and artificially inflated model performance in a basin-lumped modelling framework. In addition, when two gauges exhibit substantial overlap of their \revr{drainage} areas but provide conflicting discharge records---for instance\revr{,} due to inconsistent rating curves---a data-driven model cannot \revr{i}dentify which record is correct and instead learns an interpolated behaviour that does not accurately represent either catchment.

To address this issue, a systematic deduplication and quality-control procedure is applied based on river gauge location along the river network, the size and shape of the \revr{drainage} area, and observed discharge similarity.

Basin boundaries are taken directly from the original catchment polygon shapefiles provided by each source dataset and are reprojected to a common geographic coordinate system (EPSG:4326). Spatial overlap is computed directly from the catchment boundary polygons. Every possible pair of basin polygons is compared: for each pair, the fractional overlap is defined as the intersecting area normalised by the smaller of the two basin areas. Basin pairs with a fractional overlap greater than or equal to 0.7 are flagged for further inspection.

For each flagged pair, the Kling--Gupta Efficiency (KGE) \cite{gupta2009decomposition} is computed between the corresponding observed streamflow time series. Pairs with $\mathrm{KGE} \geq 0.95$ are classified as strict duplicates, and 2{,}007 stations are removed from the training set. Among duplicate pairs, stations with the longest observational records after 2016 are preferentially retained in order to maximize overlap with the availability of high-resolution operational forcing data. Deterministic IFS forecasts at approximately 0.1$^\circ$ horizontal resolution become available in March 2016 with the implementation of Cycle 41r2 \cite{haiden2016evaluation} and provide the high-fidelity inputs required for model fine-tuning and inference. Stations pairs with poor agreement ($\mathrm{KGE} < 0.6$) are discarded entirely, as such discrepancies likely indicate data quality issues. Intermediate cases ($0.6 \leq \mathrm{KGE} < 0.95$) are retained, as they typically correspond to genuinely distinct hydrological behaviour in nested sub-catchments.

\begin{figure}[htbp]
    \centering
    \includegraphics[width=\textwidth]{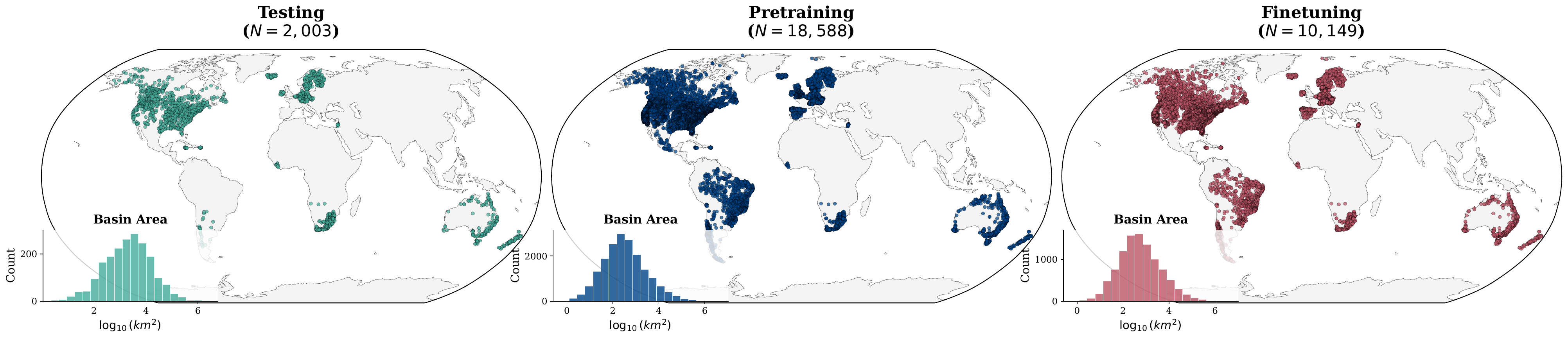}
    \caption{Global spatial distribution of the 18,588 quality-controlled streamflow stations across the three experimental stages: pre-training, fine-tuning, and testing. The inset diagrams provide the frequency distribution of basin surface areas (on a $\log_{10}$ scale) for each subset. }
    \label{fig:global_map}
\end{figure}

After an additional quality-control step that removes implausible or severely discontinuous discharge time series, a final curated set of 18{,}588 unique stations remains. In particular, non-physical or erroneous signals are identified using a flatline ratio, defined as the proportion of \revr{consecutive non-null observation pairs with identical values relative to the total number of non-null pairs}. Basins \revr{with a flatline ratio exceeding 0.9 or a discharge variance below $10^{-3}$} are excluded. The global spatial distribution of the resulting basins is shown in Figure~\ref{fig:global_map}. \revchristel{While all 18{,}588 quality-controlled basins are used during pre-training (1980--2019), subsequent phases use progressively smaller subsets dictated by data availability: 2{,}010 basins with complete records overlapping the IFS forecast period are used for fine-tuning (2016--2019), and 2{,}003 basins with continuous observations during the operational window form the temporal test set (2021--2024). }

\subsection{Model Inputs and Consistency}
The model relies on two types of inputs: dynamic features, which vary over time, and static features, which remain constant for each catchment. Dynamic features include the five core meteorological drivers: surface net solar radiation (SSR), surface net thermal radiation (STR), surface pressure (SP), 2-m air temperature (T2M), and total precipitation (TP)—as summarized in Table~\ref{tab:dynamic_inputs}. Static features consist of 203 catchment attributes describing physiography, soil properties, geology, land cover, climatology, and anthropogenic influence. These feature types are processed separately through dedicated embedding networks before being integrated into the LSTM core.

\begin{table}[ht]
    \centering
    \caption{Input features for the AIFL model. The selection is restricted to variables shared across ERA5-Land reanalysis and IFS forecasts in the \revr{Caravan} MultiMet dataset \cite{shalev2025caravan} to facilitate seamless transferability from pre-training to operational inference.}
    \label{tab:dynamic_inputs}
    \begin{tabular}{lll}
        \toprule
        Variable & Description & Unit \\
        \midrule
        SSR & Surface net solar radiation & \unit{W.m^{-2}} \\
        STR & Surface net thermal radiation & \unit{W.m^{-2}} \\
        SP  & Surface pressure & \unit{kPa} \\
        T2M & Air temperature at 2 metres & \unit{^\circ C} \\
        TP  & Total daily precipitation & \unit{mm.d^{-1}} \\
        \bottomrule
    \end{tabular}
\end{table}

While incorporating additional predictors, higher temporal resolution, or more descriptive spatial aggregations (e.g., basin-scale maxima, variability, or other statistics rather than simple catchment-averaged means) would likely improve performance, AIFL is currently constrained to this standardised daily input set. These variables are provided consistently across both ERA5-Land reanalysis and operational IFS forecasts through the \revr{Caravan} MultiMet extension \cite{shalev2025caravan}, ensuring that the input space remains identical between pre-training and operational fine-tuning.

The static catchment attributes are sourced from the \revr{Caravan} dataset \cite{kratzert2023caravan}, which offers a globally consistent description of landscape characteristics governing hydrological response. These features, primarily aggregated from HydroATLAS \cite{linke2019global}, encompass diverse environmental categories. They describe the physical structure of the basin through metrics such as mean elevation and slope, the subsurface environment via soil texture and lithology, and the surface conditions through vegetation indices and land-use fractions. Climatological indices representing long-term averages and anthropogenic descriptors, such as population density and degree of regulation, are also included to allow the model to distinguish between different hydrological regimes. 

Beyond the original \revr{Caravan} attributes, additional time-aware features are introduced to improve temporal representation. Seasonal information is encoded using sine and cosine transformations of day-of-year and month. In addition, each station’s Coordinated Universal Time (UTC) offset is provided as an explicit input to facilitate correct temporal alignment between UTC-based meteorological forcings and local-time streamflow observations. We deliberately avoid a direct conversion of streamflow time series to UTC, as the required temporal interpolation would introduce non-physical artifacts, such as the artificial smoothing of hydrograph peaks and synthetic temporal shifts, potentially degrading the integrity of the original peak-flow observations. By providing the UTC offset as a static attribute, we allow the LSTM core to account for day-boundary misalignments---where a UTC-based forcing event may correspond to the previous or following calendar day in local time---without altering the raw observational data.

\subsection{Data Availability for Temporal Evaluation}
\label{study_basin}

This study focuses on temporal generalisation, evaluating the model’s ability to forecast future events at established gauging stations. While spatial extrapolation to ungauged basins remains a critical challenge, the present experimental design specifically targets an operational forecasting setting where predictions are issued for fixed, historically observed locations. We therefore utilise the full set of stations as a consistent reference for the pre-training, fine-tuning and testing phases, ensuring the model captures a global diversity of hydrological signatures.

\begin{figure}[htbp]
    \centering
    \includegraphics[width=0.9\textwidth]{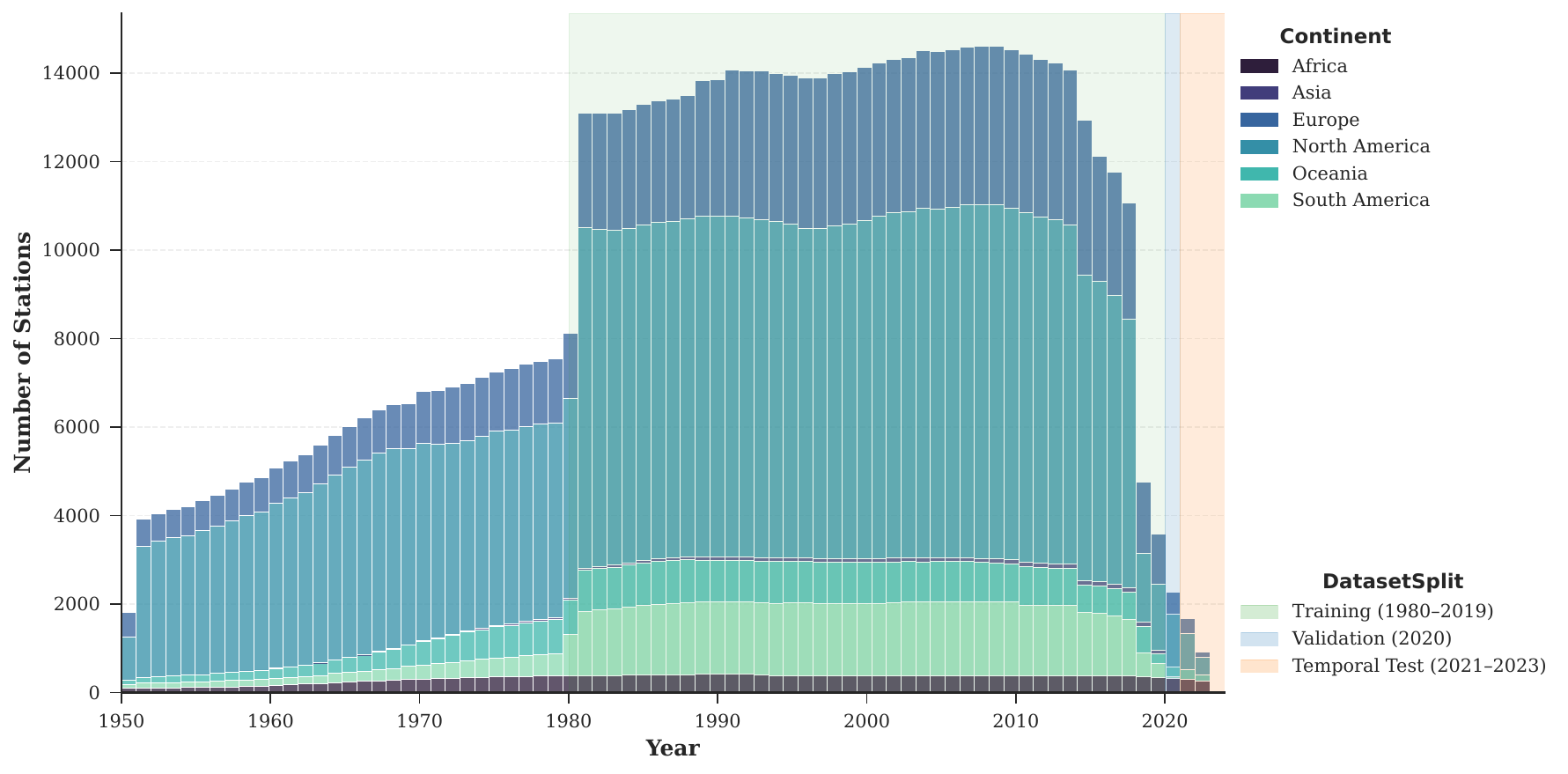}
    \caption{Global station availability over time (1950--2023). Shaded regions indicate the splits for pre-training and fine-tuning (green), validation (blue), and testing (orange).}
    \label{fig:availability}
\end{figure}

The decision to maintain a consistent set of gauging stations, rather than implementing a spatial split, is driven primarily by data availability. As shown in Figure~\ref{fig:availability}, there is a sharp decline in active station records after 2015, as many constituent datasets within the \revr{Caravan} framework lack updates beyond 2018--2019. Partitioning the remaining active stations for both spatial and temporal validation would yield a test set for the 2021--2024 period that is too small to be climatically or geographically representative.

For the independent temporal test period (2021--2024), we identified a subset of 2,003 basins where continuous streamflow observations remained available during the operational forcing window. While the overall dataset spans a consistent range of spatial scales—from small headwater catchments (approximately 1~km$^{2}$) to continental-scale river systems exceeding $10^{6}$~km$^{2}$—the subset of basins available for evaluation during the operational testing period is skewed toward larger catchments. This shift reflects the availability of streamflow gauge records that overlap with the operational forcing period and results in model evaluation being concentrated on larger river systems. Despite this inherent weighting, the 2,003-station subset maintains sufficient hydroclimatic diversity to provide a robust assessment of model performance under strictly future meteorological conditions. While developing strategies to better leverage the sparse records available for operational testing—and balancing evaluation across all spatial scales—remains a critical frontier for global hydrology, such methodological refinements are reserved for future work.

\externaldocument{04_results}

\section{Methodology}
\label{sec:methodology}

This section describes the architecture and training strategy of the AIFL model. The model is implemented using the open-source NeuralHydrology framework \cite{kratzert2022joss} as a starting point. While the core LSTM architecture follows NeuralHydrology’s implementation, we introduce several extensions and adaptations—particularly in data preprocessing, training configuration, and forecast-time inference—to support global-scale pre-training on ERA5-Land and subsequent fine-tuning with IFS operational forecasts. The methodology is designed to learn physically meaningful rainfall–runoff relationships from reanalysis data while enabling robust adaptation to operational numerical weather prediction forcings.

\subsection{Model Architecture}

The AIFL model employs a single-layer LSTM network to capture temporal rainfall–runoff dynamics across a 180-day window. This duration was chosen via sensitivity analysis to balance hydrological memory with computational efficiency; extending the window to 365 days (as seen in \cite{nearing2024global}) yielded negligible accuracy gains while significantly increasing \revr{pre-training} overhead. To handle high-dimensional inputs, dynamic features and static features are first transformed through separate three-layer feedforward embedding networks (layer sizes 30, 20, 64) with $\tanh$ activation functions. The embeddings are then concatenated and fed into the LSTM to produce 10-day output sequences. The architectural specifications are summarized in Table~\ref{tab:architecture_summary}.

\begin{table}[ht]
    \centering
    \caption{Summary of the AIFL model architecture and hyperparameter choices.}
    \label{tab:architecture_summary}
    \begin{tabular}{ll}
        \toprule
        Component & Configuration \\
        \midrule
        Recurrent layer & Single-layer LSTM \\
        Hidden state size & 1024 \\
        Dropout (output) & 0.4 \\
        Static embedding & 3-layer MLP (30, 20, 64) with $\tanh$ \\
        Dynamic embedding & 3-layer MLP (30, 20, 64) with $\tanh$ \\
        Temporal window & 180 days \\
        Output sequence length & 10 days \\
        Input features & 5 dynamic variables + 203 static attributes \\
        \bottomrule
    \end{tabular}
\end{table}

This configuration was selected based on early architectural development that evaluated generalisation performance across both temporal and spatial settings. Although the operational focus of AIFL is temporal forecasting, spatial validation was utilised to assess whether the model could learn robust physical representations of hydrological heterogeneity. 

While no systematic hyperparameter search was performed, a 1,024-unit model and a smaller 256-unit model, proposed by Ryd and Nearing~\cite{ryd2025fine}, were compared. \rev{Both architectures showed comparable skill in temporal test sets, achieving median KGE$'$ of 0.663 vs 0.641, respectively}. \rev{The larger hidden model demonstrated modestly higher representational capacity also in spatial generalization, though the limited sample of 50 hold-out basins precludes definitive conclusions regarding the impact of model size on spatial transferability. A systematic investigation into how model capacity influences the trade-off between spatial and temporal generalization is beyond the scope of this work, which focuses primarily on temporal forecasting at gauged locations. Nevertheless, the larger hidden state was retained due to its advantage in temporal performance, though we acknowledge that these improvements may be attributed to a higher capacity for memorizing basin-specific characteristics.}

\subsection{Training Strategy} \label{subsec:training}

The model is trained using a normalised Mean Squared Error (MSE) loss function following \cite{kratzert2019hess}, which ensures that all basins contribute equally to the optimization process regardless of their absolute flow magnitude:
\begin{equation}
\mathcal{L}_{\text{norm-MSE}} = \frac{1}{N} \sum_{i=1}^{N} \left( \frac{(y_i - \hat{y}_i)^2}{\sigma_{\text{basin}(i)}^2 + \varepsilon} \right),
\label{eq:nse_loss}
\end{equation}
where $y_i$ and $\hat{y}_i$ denote observed and predicted specific discharge, and $\sigma_{\text{basin}(i)}$ is the standard deviation of the observed discharge for the corresponding basin calculated over the training period. The constant $\varepsilon = 0.1$ is added for numerical stability to prevent division by very small variances, and $N$ is the total number of training samples (time steps across all basins) in the current training batch. This local normalisation of the loss function ensures that the model is not biased toward basins with high-flow variability. All dynamic and static input features are z-score normalised on a per-variable basis using global training statistics before being fed into the embedding networks.

\begin{figure}[htbp]
    \centering
    \includegraphics[width=0.8\textwidth]{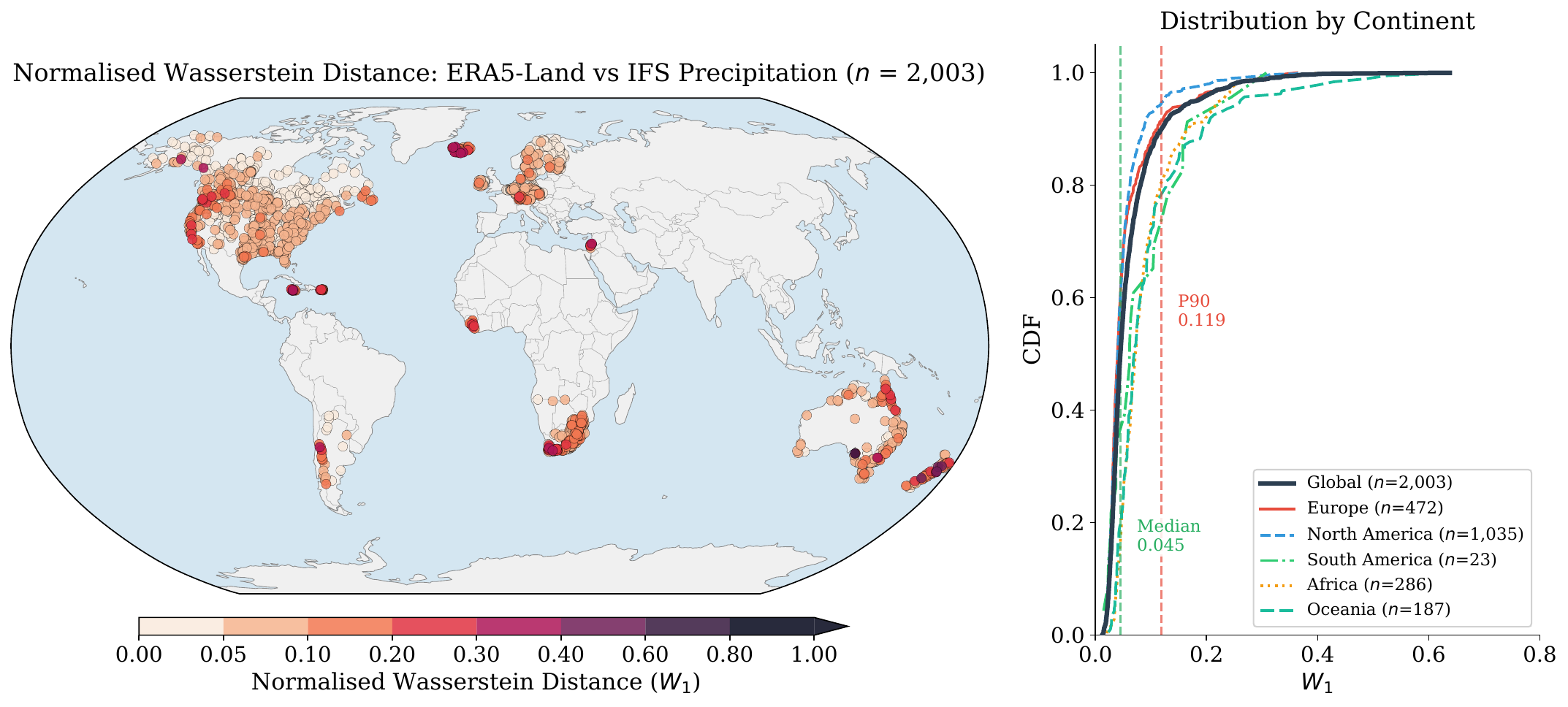}
        \caption{Global spatial distribution of the normalised Wasserstein distance ($W_1$) between ERA5-Land reanalysis and 1-day lead time (LT1) IFS daily precipitation, computed for 2{,}003 basins over a common 4-year period (2016--2019).  \rev{$W_1$ is the Earth Mover's Distance between the wet-day ($>1$\,mm) precipitation distributions, normalised by the mean ERA5-Land wet-day precipitation, yielding a dimensionless measure of relative distributional shift.} \revchristel{The right panel shows the cumulative distribution of $W_1$ by continent.}}
    \label{fig:wasserstein_map}
\end{figure}

\begin{figure}[htbp]
    \centering
    \includegraphics[width=\textwidth]{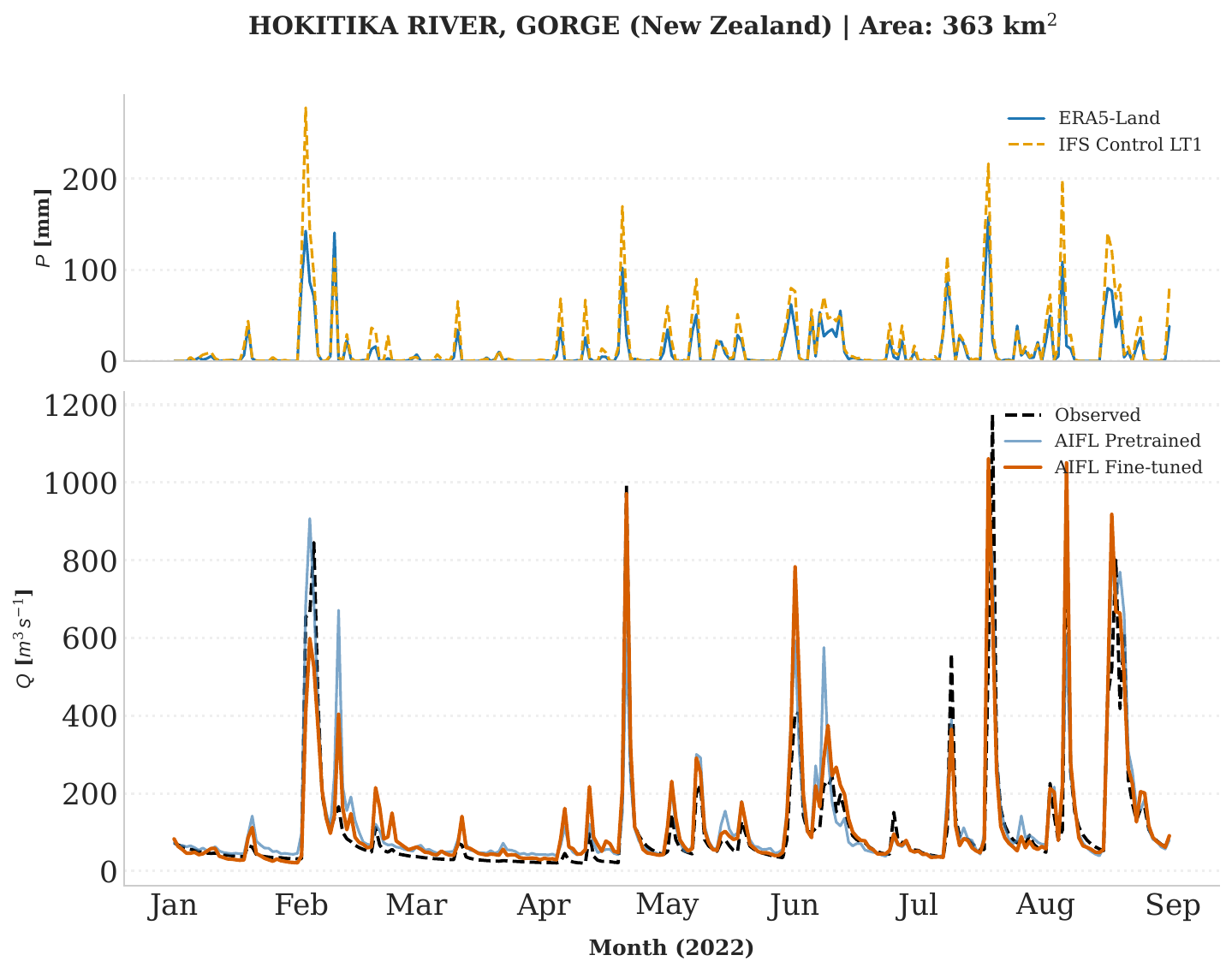}
    \caption{Hydro-meteorological time series for the Hokitika River, Gorge (New Zealand; 363 km$^2$) during 2022. Top: precipitation ($P$) from ERA5-Land (blue) and IFS Control LT1 (yellow). Bottom: observed discharge ($Q$; black dashed) compared against two AIFL model configurations both forced by the same IFS Control LT1 inputs: the pre-trained model (steel blue) and the fine-tuned model (red). This direct comparison isolates the impact of the model weights, demonstrating how the fine-tuned AIFL learns to correct for the systematic wet bias in IFS precipitation peaks to align streamflow magnitudes with observations.}
    \label{fig:hokitika_bias}
\end{figure}

Systematic distributional shifts exist between the ERA5-Land reanalysis used during model development and the operational IFS forcings encountered in real-time forecasting. To accommodate these discrepancies between historical training data and operational environments, we adopt a two-stage transfer learning strategy that transitions from reanalysis-driven pre-training to forecast-aligned fine-tuning. Unlike the coupled data assimilation used in operational IFS streams, ERA5-Land is produced as an offline land-surface "replay" with specific elevation corrections for thermodynamic states, resulting in a distinct statistical signature \cite{munoz2021era5}.   As illustrated by the normalised Wasserstein distance in Figure~\ref{fig:wasserstein_map}, discrepancies between ERA5-Land and operational IFS precipitation can be substantial across large parts of the globe. \revchristel{Quantitatively, the global median $W_1$ is small (0.045, i.e., 4.5\% of mean wet-day precipitation), while the upper decile exceeds 0.119, with extreme cases reaching 0.638.} Geographically, North America and Europe exhibit the most consistent distributions, while Africa and South America show wider distributional tails, representing regions where the operational IFS precipitation forcings deviate most significantly from ERA5-Land. Without an explicit alignment phase, a model trained solely on reanalysis would propagate these forcing inconsistencies directly into its streamflow simulations, leading to degraded performance in real-time deployment. Our strategy resolves this by first utilizing the multi-decadal, physically consistent ERA5-Land record (1980–2019) to learn universal hydrological response functions, before subsequently fine-tuning all model weights on IFS control forecasts (2016–2019). This second stage allows the LSTM to adapt its internal representations to the specific error structures and statistical characteristics of the operational NWP stream, effectively correcting for forecast-induced biases like those observed in the Hokitika River example (Figure~\ref{fig:hokitika_bias}). 

In the pre-training phase, the network is optimised on ERA5-Land reanalysis inputs using 180-day sequences spanning 1 January 1980 to 31 December 2019 across all 18,588 basins (Table~\ref{tab:experimental_phases}). The objective of this phase is to learn general, physically meaningful rainfall–runoff relationships from a long and internally consistent historical record. Training uses the Adam optimizer with an initial learning rate of $4 \times 10^{-4}$, gradient clipping at 1.0, and Gaussian target noise ($\sigma = 0.02$) to improve robustness. A cosine annealing learning-rate schedule is applied over 100 epochs, including 10 warm-up epochs\rev{, during which the learning rate is linearly ramped from $0.1\times$ the target rate to the full value to avoid destabilising the pretrained weights}. Each epoch is capped at 10,000 parameter updates with a batch size of 512. Validation is performed every 10 epochs on a fixed subset of 1,000 basins to monitor convergence and prevent overfitting.

Following pre-training, the model is fine-tuned using IFS control forecasts with a 1-day lead time (LT1) over the 2016–2019 period. \rev{All 180 input time steps use LT1 forcing: each day in the lookback window uses the IFS forecast issued on that day for a 1-day horizon, so the model sees 180 consecutive daily LT1 forecasts from 180 separate issuances, rather than a single forecast trajectory with increasing lead times.} This second phase explicitly aligns the learned hydrological representation with the statistical properties and systematic biases of the operational forcing. \revtwo{Using LT1 forcing uniformly across the full input sequence, rather than only at the prediction horizon, is a deliberate choice. It keeps the entire sequence within a single, internally consistent ECMWF data ecosystem (ERA5-Land for pre-training and IFS for fine-tuning) and avoids the distribution discontinuity that mixing data sources within a single sequence would introduce at the lookback--forecast boundary.} We specifically utilise LT1 for fine-tuning and validation as it provides the most direct representation of the operational IFS data distribution while minimizing the compounding atmospheric errors associated with longer lead times. By aligning the model with the physics and resolution of the LT1 stream, we establish a robust baseline for operational performance; while the operational system extends to a 10-day horizon, the primary objective of this stage is to correct for the fundamental reanalysis-to-forecast shift rather than lead-time-specific drift. \revtwo{The use of lead-time-dependent forcing, in which each step of the input window would draw on a forecast issued at the corresponding lead time, is reserved for future analyses.} \rev{To validate the two-stage strategy, we conducted ablation experiments comparing: (a)~a naive baseline trained on IFS forcing only (no ERA5-Land pretraining), and (b)~a mixed-forcing one-stage model trained with random ERA5-Land/IFS swapping. All models were trained to convergence, with cosine-annealed learning rates reach near-zero by the final epoch and training losses flat over the last 10--15\% of training. Over 2{,}010 test basins, the two-stage model achieves a median KGE$'$ of 0.663, compared to 0.645 for the naive baseline and 0.604 for the mixed-forcing approach. The two-stage model outperforms the naive baseline in 56.3\% of basins and the mixed-forcing model in 64.0\%, with the largest gains concentrated in the performance tail (mean $\Delta$KGE$' = +0.084$ vs naive). Importantly, while this ablation focuses on the 2{,}010 basins common to both forcing datasets, the pretraining stage leverages the full set of 18{,}588 basins with available ERA5-Land forcing data. We therefore attribute the two-stage advantage to the transfer of knowledge from this significantly larger global sample. By drawing on thousands of additional catchments, the pretraining stage broadens the model's representation of global hydrological diversity; this not only benefits the fine-tuning basins but also improves the model's spatial coverage and generalization potential in regions where IFS-specific training data is absent.} 

To ensure numerical continuity, the ERA5-Land-based input scaler is reused and applied to IFS variables through a one-to-one variable mapping. \rev{Similarly, the per-basin standard deviations ($\sigma_{\text{basin}}$) used in the normalised MSE loss (Equation~\ref{eq:nse_loss}) are inherited from the pre-training period (1980--2019), preventing an abrupt shift in loss normalisation between training stages.} All model parameters remain unfrozen, allowing the network to adapt its internal representations to forecast-specific error structures. Fine-tuning uses the same sequence length and prediction horizon as pre-training, but with a reduced initial learning rate of $1 \times 10^{-4}$ over 30 epochs and five warm-up epochs. This lower learning rate limits catastrophic forgetting of the physical relationships learned during pre-training while providing sufficient flexibility to correct forecast-induced biases.  \rev{We verified training stability in two ways. First, repeating only the fine-tuning stage with three different random seeds (all starting from the same pretrained checkpoint) yields median KGE$'$ values of 0.661--0.665, with a median inter-seed standard deviation of 0.022 per basin, confirming that fine-tuning is highly reproducible. Second, repeating both pretraining and fine-tuning end-to-end with three different seeds produces median KGE$'$ values of 0.660--0.674, with a median inter-seed standard deviation of 0.051. The wider spread indicates that pretraining initialisation contributes more variability than the fine-tuning seed alone, though all reruns remain within $\pm$0.01 of the selected model's median KGE$'$ of 0.663.}

\begin{table}[ht]
    \centering
    \caption{AIFL experimental design and operational periods across the 18,588-station network.}
    \label{tab:experimental_phases}
    \begin{tabular}{llll}
        \toprule
        Phase & Input Data Source & Period & Primary Purpose \\
        \midrule
        Pre-training & ERA5-Land Reanalysis & 1980 -- 2019 & Learn universal hydrology \\
        Fine-tuning & IFS Control (LT1) & 2016 -- 2019 & Operational alignment \\
        Validation  & IFS Control (LT1) & 2020 & In-time generalisation \\
        Temporal Test & IFS Control (LT1) & 2021 -- 2024 & Operational performance \\
        \bottomrule
    \end{tabular}
\end{table}

The transition from reanalysis-driven pre-training to forecast-aligned fine-tuning substantially reshapes the performance distribution across the 2{,}003 basins in the temporal test set. While the medians remain largely stable—with a median $\Delta \mathrm{KGE}'$ (the modified Kling–Gupta Efficiency; \cite{kling2012runoff}) of $-0.013$ and a median $\Delta \mathrm{NSE}$ of $-0.032$—the two-stage training strategy leads to a pronounced improvement in mean global skill. Specifically, the mean $\mathrm{KGE}'$ increases from 0.21 to 0.44, while the mean NSE improves from $-11.40$ to $-3.26$.

The divergence between median and mean changes indicates that fine-tuning primarily improves previously low-performing basins, contracting the lower tail of the performance distribution where the pre-trained model fails to generalise to operational forcings. Overall, $\mathrm{KGE}'$ improves in 44.7\% of basins, with 17.7\% of the full sample exhibiting gains exceeding 0.1. Figure~\ref{fig:hokitika_bias} illustrates this corrective behaviour in the Hokitika River basin. The pre-trained model exhibits a persistent positive bias, consistently overestimating baseflow and recession levels compared to observations. Fine-tuning effectively corrects this systematic volume error to better align with observed discharge.

The response to fine-tuning is spatially non-uniform. While 22.7\% of basins experience notable declines in performance ($\Delta \mathrm{KGE}' < -0.1$, defined as the change in $\mathrm{KGE}'$ between the fine-tuned and pre-trained models), these are outweighed by improvements in previously poorly performing catchments, where the mean increase in $\mathrm{KGE}'$ among significantly improved basins is 1.75.

This asymmetry suggests that fine-tuning acts primarily as an operational stabiliser: it mitigates severe forecast-induced errors at the cost of minor degradations in already well-calibrated basins. The operational significance of this trade-off depends on the relative importance of the affected basins, and future work could explore selective or basin-weighted fine-tuning strategies. Together, these results motivate the explicit separation between pre-training on internally consistent reanalysis data and forecast-aligned fine-tuning as a practical strategy for managing distributional drift between reanalysis and operational meteorological forcing streams.

\section{Results and Evaluation}
\label{sec:results}

This section evaluates the predictive skill of the AIFL model with a focus on temporal generalisation, operational flood forecasting performance, and comparison against current global benchmarks. All results are reported for a strictly temporal test setting, assessing the model’s ability to forecast future conditions at previously observed gauging locations.

\subsection{Temporal generalisation and Global Performance}

AIFL is evaluated over an independent temporal test period spanning 1 January 2021 to 30 September 2024 at 2,003 gauged basins for which continuous streamflow observations are available during the operational forcing period. Model predictions are compared against observed daily streamflow to assess predictive skill at known locations under operational conditions. The test basins represent a geographically diverse subset of the curated dataset and were selected solely based on the availability of observations overlapping with the operational IFS forecast period, rather than through spatial sampling, as detailed in Section~\ref{study_basin}.

\rev{All results reported in this section correspond to forecasts driven by IFS Control precipitation at LT1. All 180 input steps and all 10 output predictions therefore use LT1-quality forcing; metrics are computed over the merged chronological predictions from overlapping evaluation windows.} Across the test basins, AIFL achieves a median $\text{KGE}'$ of 0.66. For context, the GloFAS v4 operational system reports a median $\text{KGE}'$ of 0.70 for its 1,995 calibrated stations when evaluated against ERA5-forced reanalysis \cite{grimaldi2023glofas}. A detailed performance comparison between AIFL and the Google global flood model \cite{nearing2024global}, conducted across a shared subset of gauging stations, is provided in Section~\ref{subsec:benchmarking_google}.

\begin{figure}[htbp]
\centering
\includegraphics[width=\textwidth]{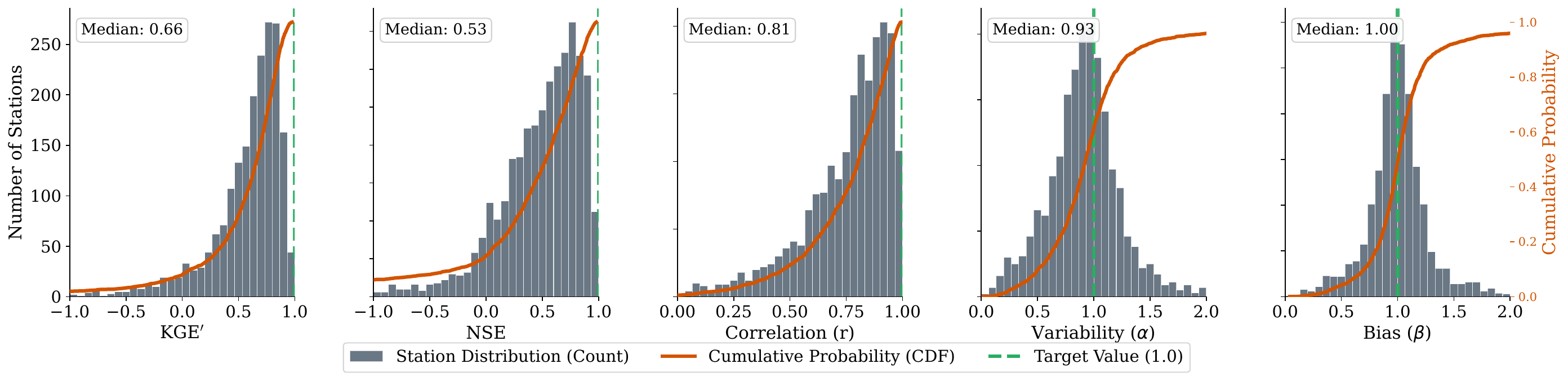}
\caption{\rev{Global performance metrics for the AIFL model across 2,003 test basins. Grey histograms show the station frequency distribution and red curves the empirical cumulative distribution function (CDF). Green vertical dashed lines denote the ideal target value (1.0) for each metric; median values are annotated in each panel. From left to right: KGE$'$ (median 0.66), NSE (0.53), Pearson correlation $r$ (0.81), variability ratio $\alpha$ (0.93), and bias ratio $\beta$ (1.00).}}

\label{fig:metric_distributions}
\end{figure}

The observed skill reflects the model's ability to reproduce streamflow dynamics accurately while correcting for systematic biases in the forcing data. The model maintains a high median Pearson correlation ($r = 0.81$), demonstrating that the LSTM reliably captures the timing and temporal structure of streamflow, while the two-stage fine-tuning strategy mitigates the systematic ``wet'' bias typical of raw operational forecasts, achieving a median bias ratio ($\beta$) of 1.00. Metric decomposition (Figure~\ref{fig:metric_distributions}) confirms that these improvements arise from a combination of strong temporal agreement and neutralization of systematic forcing errors. With a median NSE of 0.53\rev{---noting that approximately half the stations fall below this value, predominantly in arid and intermittent-flow basins where NSE is particularly sensitive to mean-flow bias---the model demonstrates robust temporal generalisation at gauged locations while producing reliable streamflow predictions with near-unbiased volume estimates}.

\begin{figure}[htbp]
\centering
\includegraphics[width=\textwidth]{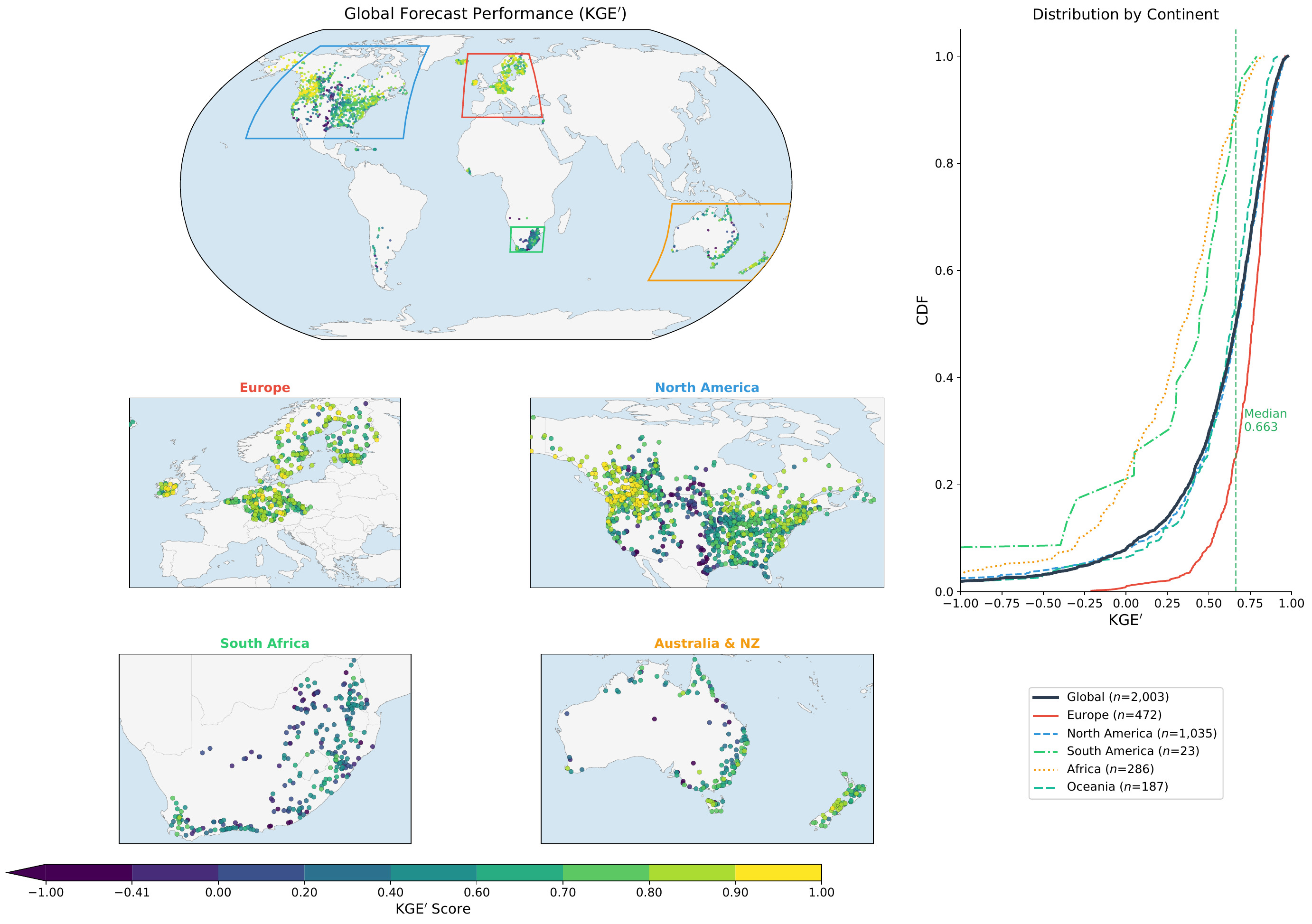}
\caption{\rev{Global and regional forecast performance (KGE$'$) for the 2021--2024 test period. The left panels show the global distribution of station-level KGE$'$ (top), and detailed regional insets for Europe, North America, South Africa, and Australia/New Zealand (bottom). The right panel shows the cumulative distribution of KGE$'$ by continent: North America and Europe achieve the highest median skill, while Africa shows the widest spread and lowest median, consistent with the concentration of arid and intermittent-flow basins in that region.}}
\label{fig:map_kge_prime}
\end{figure}

The spatial distribution of KGE$'$ values is shown in Figure~\ref{fig:map_kge_prime}. Skill is consistently high across Europe, North America, and Oceania, with most basins achieving KGE$' > 0.6$. Lower performance is concentrated in arid and semi-arid regions and in basins characterised by highly-intermittent flow regimes. These catchments typically exhibit long periods of near-zero discharge punctuated by sparse, sharp peaks, which strongly penalize composite performance metrics such as KGE$'$.

\begin{figure}[htbp]
\centering
\includegraphics[width=\textwidth]{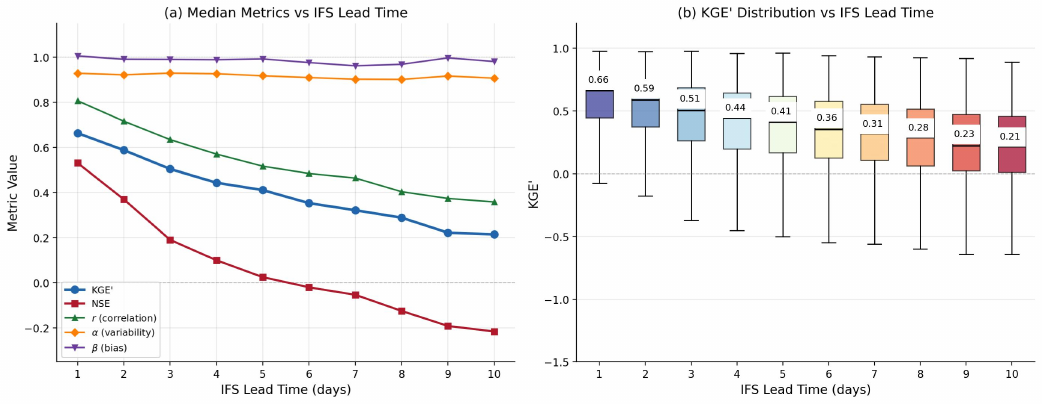}
\caption{\rev{AIFL performance as a function of IFS forcing lead time calculated across 2{,}003 test basins. (a)~Median KGE$'$, NSE, and Pearson~$r$ decrease monotonically from LT1 to LT10. NSE becomes negative beyond LT5, indicating the model performs worse than the mean-flow baseline at longer lead times. (b)~Distribution of KGE$'$ for each lead time. The same model weights are used throughout; only the input forcing quality changes.}}
\label{fig:lead_time_degradation}
\end{figure}

\rev{To quantify the sensitivity to forcing quality, we evaluated AIFL with IFS forecasts at lead times 1 through 10. The same model weights are utilized, and only the input forcing changes. Figure~\ref{fig:lead_time_degradation} shows that performance degrades monotonically: median KGE$'$ drops from 0.663 (LT1) to 0.215 (LT10), and median NSE becomes negative beyond LT5. This confirms that the reported LT1 metrics represent the model's best achievable skill and that extending the forecast horizon with degraded atmospheric forcing significantly erodes hydrological prediction quality. Notably, this degradation is almost entirely driven by a reduction in correlation, whereas the variability  and bias  components remain relatively stable across all lead times. This suggests that while the model maintains the correct physical magnitude and average volume of streamflow, its predictive skill is highly sensitive to the temporal misalignment of the longer-lead meteorological forcings}. 


\subsection{Flood Event Performance}
\label{subsec:flood_event_perf}

Beyond overall temporal accuracy, operational flood forecasting critically depends on the reliable detection of high-flow events. To evaluate this, we define flood events relative to return-period thresholds derived from a long-term reference simulation. Specifically, the AIFL model is driven by ERA5-Land inputs over 1980--2024 to generate multi-decadal simulated discharge time series for each basin. Annual maxima extracted from these simulations are used to estimate return-period thresholds ranging from 1.5 to 50 years using a standard extreme-value framework.  While the framework can theoretically produce higher-magnitude estimates, we limit our analysis to a 50-year maximum to ensure statistical robustness. These thresholds are derived by fitting a Gumbel distribution using the first two L-moments ($\lambda_1, \lambda_2$) to provide stable estimates of the distribution's location and scale parameters \cite{hosking1997regional}. Defining thresholds in simulated discharge space ensures statistical consistency with the model outputs and avoids biases that would arise from applying observational thresholds, particularly in regions with sparse or uncertain discharge records. To maintain this consistency during evaluation, we adopt a dual-threshold framework: AIFL forecast events are defined by exceedances of thresholds derived from the 45-year (1980--2024) ERA5-Land historical simulation, while observed events are identified when gauge discharge exceeds thresholds derived from the corresponding historical observational record. This utilization of the full climatological record maximizes the sample size of the annual maxima series, providing more stable Gumbel distribution parameters for return periods up to 50 years.

Flood event performance is quantified using \rev{three standard verification metrics. The probability of detection (POD), also known as recall, measures the fraction of observed flood events that are correctly identified by the model, indicating sensitivity to true extremes. The false alarm ratio (FAR) measures the fraction of predicted flood events that do not correspond to an observed exceedance, quantifying the rate of spurious warnings. Precision, defined as $1 - \text{FAR}$, gives the complementary view: the fraction of predicted events that coincide with genuine observed exceedances. These metrics are computed exclusively for the predefined test basins under a strict zero-day timing criterion, requiring exact-day coincidence between predicted and observed threshold exceedances with no allowed temporal lag.}

\rev{When evaluated using exact day-to-day event matching under the dual-threshold framework, where model predictions are compared against model-specific Gumbel return levels and observations against observation-derived thresholds, the model achieves high precision (0.85--1.00) with a FAR of 0.15 or below for return periods up to 5 years, dropping to zero for rarer events (Table~\ref{tab:moe0_event_stats}). This is consistent with the model's conservative tendency to underestimate peak magnitudes: exceedances in the model's own climatological reference frame are rare and, when they occur, reliably correspond to genuine observed events. Such behaviour is valuable for operational early-warning systems where false positives erode user trust and trigger unnecessary mobilisation of emergency resources~\cite{barnes2007, schroter2017}.}

\rev{However, this high dual-threshold precision is partly an artifact of the framework itself. When a model exhibits systematic positive bias, its fitted return levels are elevated relative to observed thresholds, mechanically suppressing false alarms. To expose the true false alarm behaviour, we additionally evaluate under a single-threshold framework where both predicted and observed exceedances are identified using thresholds derived from the observed annual maxima only. This reveals substantially higher FAR: 0.58 for 1.5-year events rising to 1.0 for return periods of 50 years and above, reflecting the model's tendency to overpredict extreme magnitudes relative to observed thresholds. The POD is moderate at short return periods (0.50 for 1.5-year events) but decreases with event rarity, reaching zero for return periods $\geq$\,20 years under both frameworks.}

\rev{This trade-off---moderate detection skill with substantial false alarms under the single-threshold perspective, versus high precision with limited sensitivity under dual thresholds---is consistent with typical hydrological forecast systems and provides an honest characterisation of the model's current flood detection capability. In an operational setting, the dual-threshold framework remains the appropriate verification tool, as the forecaster issues warnings based on exceedance of calibrated model return levels rather than attempting to match observed discharge magnitudes. Nevertheless, improving recall for rare extremes without inflating false alarms remains an important direction for future development.}

\begin{table}[t]
\centering
\rev{%
\caption{\rev{Global event-based verification statistics for AIFL under dual-threshold and single-threshold frameworks for the test set. \emph{Dual threshold}: observed and simulated exceedances are identified using Gumbel thresholds fitted independently to their respective annual maxima. \emph{Single threshold}: both observed and simulated exceedances are identified using thresholds fitted to the observed annual maxima only.}}
\label{tab:moe0_event_stats}
\begin{tabular}{c cc cc cc}
\toprule
 & \multicolumn{2}{c}{FAR} & \multicolumn{2}{c}{POD} & \multicolumn{2}{c}{Precision} \\
\cmidrule(lr){2-3} \cmidrule(lr){4-5} \cmidrule(lr){6-7}
Return Period & Dual & Single & Dual & Single & Dual & Single \\
\midrule
1.5\,yr  & 0.153 & 0.580 & 0.390 & 0.500 & 0.847 & 0.420 \\
2\,yr    & 0.159 & 0.587 & 0.308 & 0.457 & 0.841 & 0.413 \\
5\,yr    & 0.143 & 0.627 & 0.143 & 0.353 & 0.857 & 0.373 \\
10\,yr   & 0.000 & 0.706 & 0.000 & 0.250 & 1.000 & 0.294 \\
20\,yr   & 0.000 & 0.882 & 0.000 & 0.000 & 1.000 & 0.118 \\
50\,yr   & 0.000 & 1.000 & 0.000 & 0.000 & 1.000 & 0.000 \\
100\,yr  & 0.000 & 1.000 & 0.000 & 0.000 & 1.000 & 0.000 \\
\bottomrule
\end{tabular}
}
\end{table}

\rev{A detailed example of how AIFL is implemented in practice, including the application of the dual-threshold framework, is presented in Appendix~\ref{app:case_study}, through a case study of the Storm Henk event (December 2023 -- January 2024).}

\revtwo{A key limitation underlying this framework is that the model's simulated return levels are systematically lower than their observation-based counterparts. Across the test set, simulated Gumbel thresholds are on average approximately 90\% of the corresponding observed values at the 10- and 20-year return periods, so for most basins the underestimation is mild. A minority of basins (around 12\% of the test set), including the Storm Henk case study presented in Appendix~\ref{app:case_study}, are more strongly underestimated, with simulated return levels close to two-thirds of the observed values; for that station the simulated and observed thresholds differ by up to approximately 30\% at the 20-year level. This negative bias in extreme-value magnitudes is a direct consequence of the deterministic MSE-based loss function, which penalises large residuals quadratically and thereby suppresses peak predictions. While this underestimation does not affect warning issuance in the dual-threshold operational setting, since alerts are triggered by exceedance of the model's own calibrated levels, it does mean that forecast peak magnitudes should not be interpreted as physical discharge estimates without post-processing. Transitioning to distributional or asymmetric loss functions that better capture heavy-tailed behaviour is expected to reduce this bias and is reserved for future work.}

\subsection{Benchmarking: AIFL vs.\ Google Global Model}
\label{subsec:benchmarking_google}

The AIFL model is benchmarked against the Google global flood model \cite{nearing2024global} using publicly released outputs archived on Zenodo \cite{nearing2023zenodo}. The Google model is trained on 5,680 gauging stations, primarily from the GRDC, and employs a multi-source precipitation forcing pipeline that combines ERA5-Land and IFS reanalyses with NOAA CPC gauge data and NASA IMERG satellite estimates \cite{nearing2024global}. The evaluation considers the 1,218 stations shared between both datasets over the test period 2021--2024.

\begin{figure}[htbp]
\centering
\includegraphics[width=\textwidth]{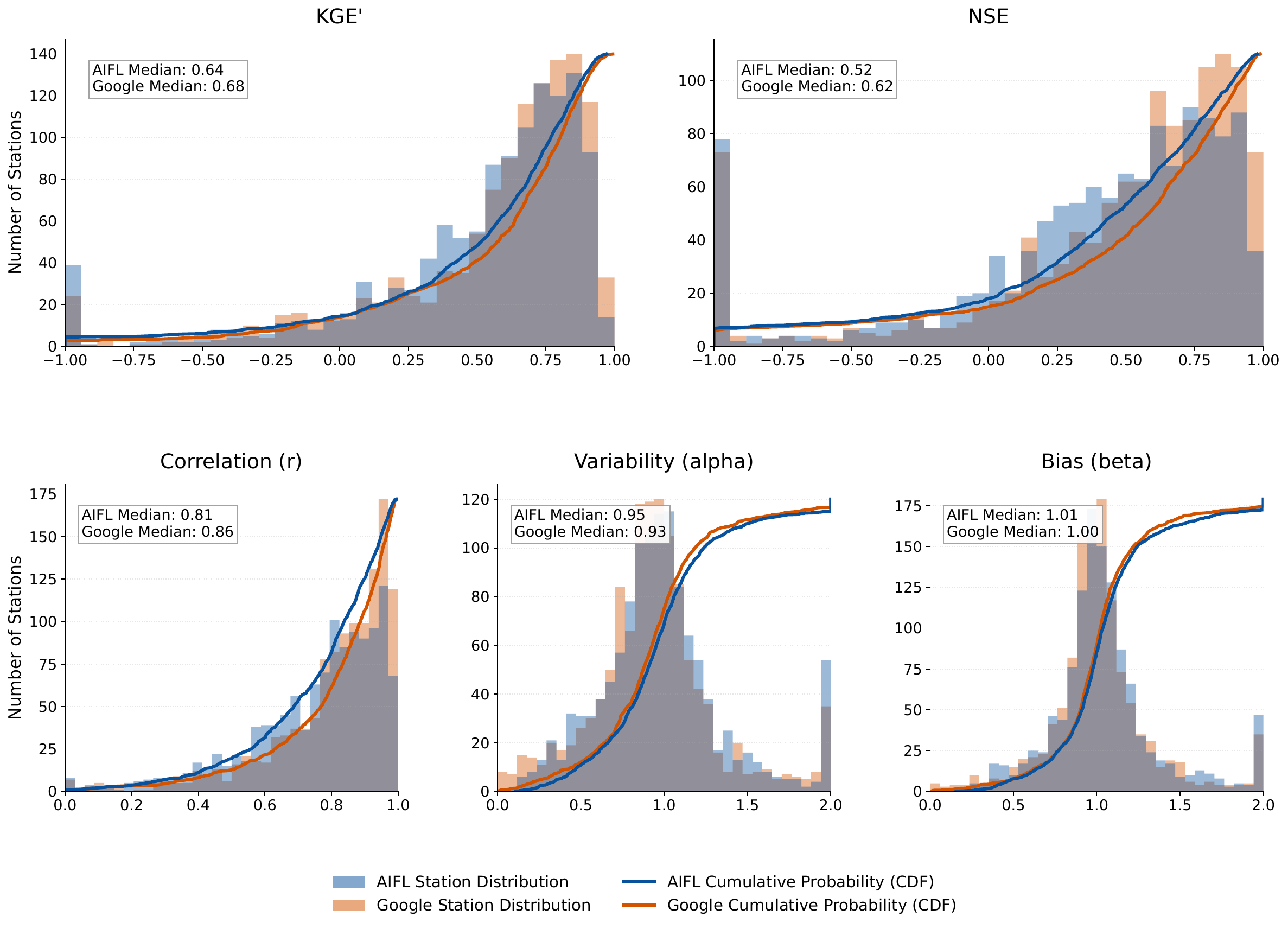}
\caption{Comparative performance metrics between AIFL (navy) and the Google global model (orange) across 1,218 shared stations. Histograms show the frequency of performance scores, while curves show empirical cumulative distribution functions (ECDFs) for KGE$'$, NSE, correlation, variability, and bias.}
\label{fig:comparison}
\end{figure}

Across this evaluation set, the Google model achieves slightly higher median skill than AIFL, with KGE$'$ values of 0.678 versus 0.636, NSE of 0.624 versus 0.518, and Pearson correlation of 0.857 versus 0.808 (Figure~\ref{fig:comparison}). Bias ($\beta$) and variability ($\gamma$) ratios are close to 1.0 for both models, indicating comparable performance in capturing long-term discharge volumes and flow variability.

\begin{figure}[htbp]
\centering
\includegraphics[width=\textwidth]{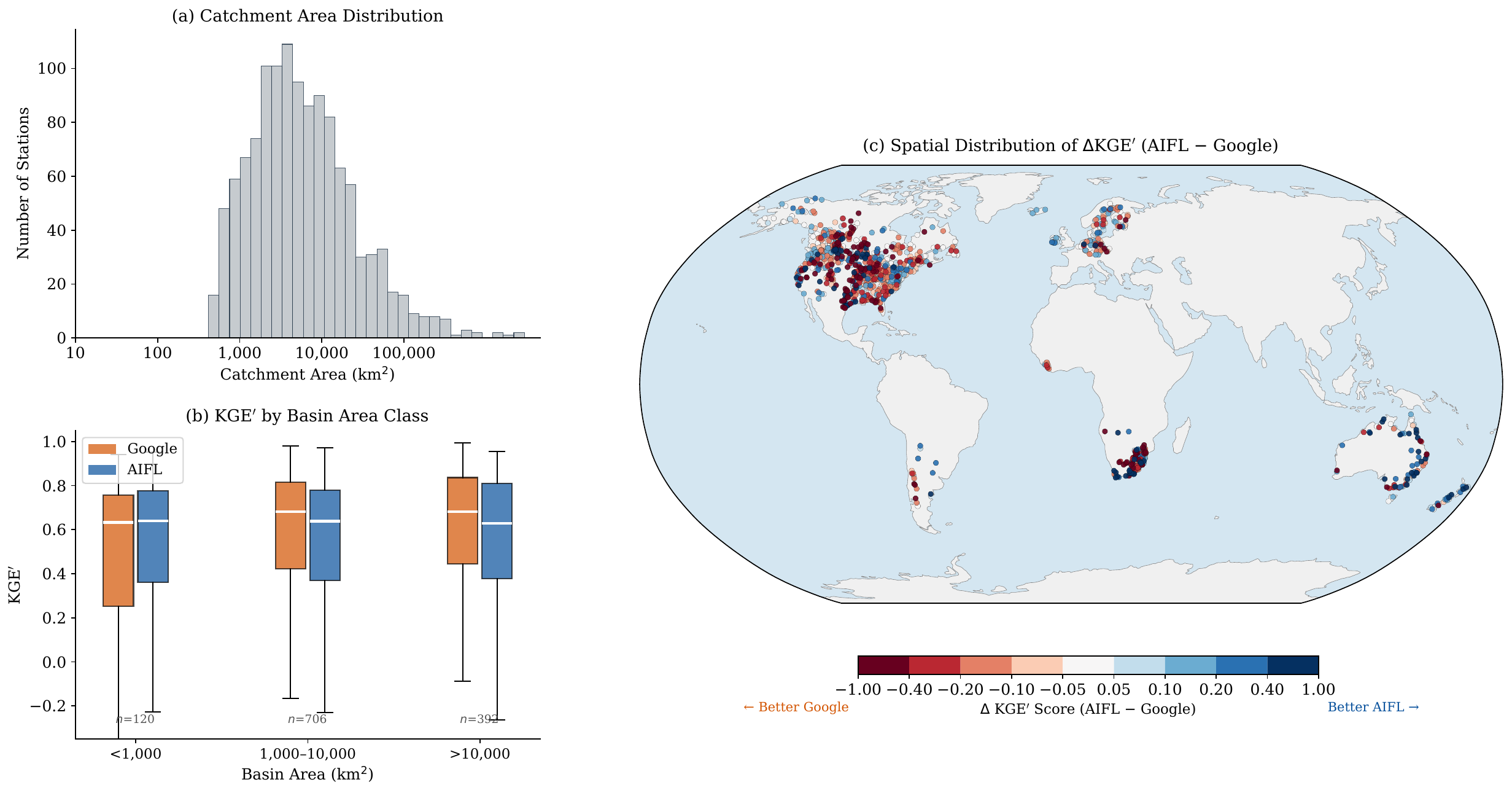}
\caption{\rev{Diagnostic benchmarking of AIFL against the Google global model across 1,218 shared evaluation stations. (a) Distribution of catchment areas on a logarithmic scale, showing that most shared stations fall in the 1,000--10,000\,km$^2$ range. (b) KGE$'$ performance stratified by basin area class ($<$1,000\,km$^2$, 1,000--10,000\,km$^2$, $>$10,000\,km$^2$), with medians and interquartile ranges for both models. (c) Spatial distribution of $\Delta$KGE$'$ (AIFL minus Google): blue markers indicate basins where AIFL achieves higher skill, red markers indicate basins where Google performs better. Markers are sorted so that stations with the largest absolute differences are plotted on top.}}
\label{fig:google_benchmarking_diagnostic}
\end{figure}

At the station level, AIFL outperforms Google at 523 locations (42.9\%), whereas Google is superior at 695 stations (57.1\%). Performance differences are most pronounced at the extremes. A subset of 125 stations (10.3\% of the shared set) exhibits very poor AIFL performance (KGE$'<0$). Of these, 44 stations (35\%) are also poorly predicted by Google, indicating that many failures occur in inherently challenging basins with complex hydrology or with systematically biased forcing data. 
In the remaining cases where AIFL skill is low, the Google model demonstrates moderate to high predictive skill. This divergence highlights specific instances where the Google model's multi-source forcing or architectural features successfully mitigate failure modes encountered by the more streamlined AIFL model. The evaluation set spans a broad range of hydrological scales, from headwater catchments under 1,000~km$^2$ to large continental basins (Figure~\ref{fig:google_benchmarking_diagnostic}a). Performance systematically varies with basin size: in smaller catchments, AIFL is highly competitive, outperforming Google at 55\% of stations, reflecting its capacity to capture rapid hydrological responses. For larger basins, the fraction of stations where AIFL underperforms rises to 62\%. Beyond median skill, AIFL demonstrates consistent performance across all basin sizes, providing a stable global baseline. While the Google model achieves slightly higher maximum skill in large basins, its performance exhibits greater variability in smaller catchments, with a wider interquartile range (IQR = 0.504), whereas AIFL maintains relatively stable skill across all basin sizes (IQR = 0.42) (Figure~\ref{fig:google_benchmarking_diagnostic}b). Spatial patterns of performance differences (Figure~\ref{fig:google_benchmarking_diagnostic}c) do not indicate strong geographic clustering, though modest regional gains are observed in Australia and Northern Europe, offset by Google’s advantages in parts of North America and Southern Africa.

The observed performance gap relative to Google can be attributed to two main factors. First, the multi-source precipitation forcing employed by Google enables the network to exploit complementary error structures across reanalysis, gauge-based, and satellite-derived products, reducing extreme failures in data-sparse or meteorologically complex regions. Second, the Google model utilises a distributional training objective based on an asymmetric Laplace loss \cite{nearing2024global}. This choice enables the model to represent a conditional distribution of discharge, providing inherent probabilistic calibration and increased robustness to outliers. In contrast, AIFL relies on a deterministic loss function with a simpler forcing pipeline. Despite these differences, AIFL achieves competitive median skill at most stations and matches or exceeds Google at over 40\% of locations, offering \rev{an independent and operationally streamlined} global baseline suitable for research, diagnostic evaluation, and deployment where simplicity and reproducibility are priorities.

\section{Conclusion and Future Directions}
\label{sec:conclusion}

This study presents AIFL, a machine learning model for global streamflow forecasting, designed to provide robust predictions under operational forcing conditions. The model is trained end-to-end using the \revr{Caravan} ecosystem, employing a two-stage strategy that first pretrains on four decades of ERA5-Land reanalysis and subsequently finetunes on operational IFS control forecasts. This approach enables adaptation to the statistical properties and biases of near-real-time forecast inputs, yielding median KGE$'$ of 0.66, NSE of 0.53 and a median bias ratio ($\beta$) of 1.00, over the 2021--2024 test period at 2,003 gauged basins. 

When evaluated against the Google global flood model at 1,218 shared stations, AIFL matches or exceeds Google skill at 42.9\% of locations. Differences are most pronounced at extremes: a subset of stations exhibits poor performance for AIFL but moderate-to-high skill for Google, highlighting basins where model-specific factors—rather than universally challenging conditions—drive errors. Performance varies systematically with basin size, with AIFL performing better in smaller catchments, while Google has an advantage in the largest basins. Beyond median skill, AIFL provides stable and consistent predictions across all basin sizes, whereas Google shows increased variability in smaller catchments.  

Event-based verification indicates that AIFL operates with a highly conservative detection profile. Under a strict zero-day timing criterion, the model achieves \revtwo{high precision (0.85--1.00) under the dual-threshold framework, with a false alarm ratio of 0.15 or below for frequent events (return periods up to 5 years) and zero false alarms for rarer events (return periods $\geq$\,10 years)}. While the model maintains high precision, its recall is limited, successfully identifying approximately half of frequent events and one-third of 50-year floods. This performance profile reflects a prioritization of high-confidence alerts; however, the high omission rate for extreme events represents a substantial trade-off. We acknowledge that for operational utility, such a configuration must be carefully balanced against the risk-tolerance of stakeholders to ensure the system remains a reliable tool for emergency response. The practical utility of this strategy was evidenced during the January 2024 Storm Henk floods in Belgium, where AIFL successfully identified a 20-year flood signal six days in advance.

Several avenues remain for future development. While this study utilised a deterministic loss function, future iterations of \revchristel{AIFL} could transition toward distributional objectives to enable inherent uncertainty quantification. \rev{By predicting a full probability distribution rather than a single point estimate, \revchristel{the model} could also better account for the heavy tails of hydrological extremes, potentially improving the probability of detection for rare events. A natural extension is to train \revchristel{AIFL} with lead-time dependent forcing inputs, explicitly accounting for IFS forecast error growth across lead times. This could improve skill beyond LT1 and provide a more realistic representation of operational forecast uncertainty at extended horizons.} Moreover, the integration of probabilistic ensemble forcing could further refine event detection and support risk-based decision-making. Future research may also investigate the impact of multi-source precipitation products on improving recall for rare extremes. By providing \rev{\revr{an independent}} architecture that balances consistency with competitive skill, AIFL establishes \rev{an operationally viable and reproducible} baseline for global streamflow forecasting.

\appendix
\section{Storm Henk Case Study}
\label{app:case_study}

\begin{figure}[htbp]
\centering
\includegraphics[width=\textwidth]{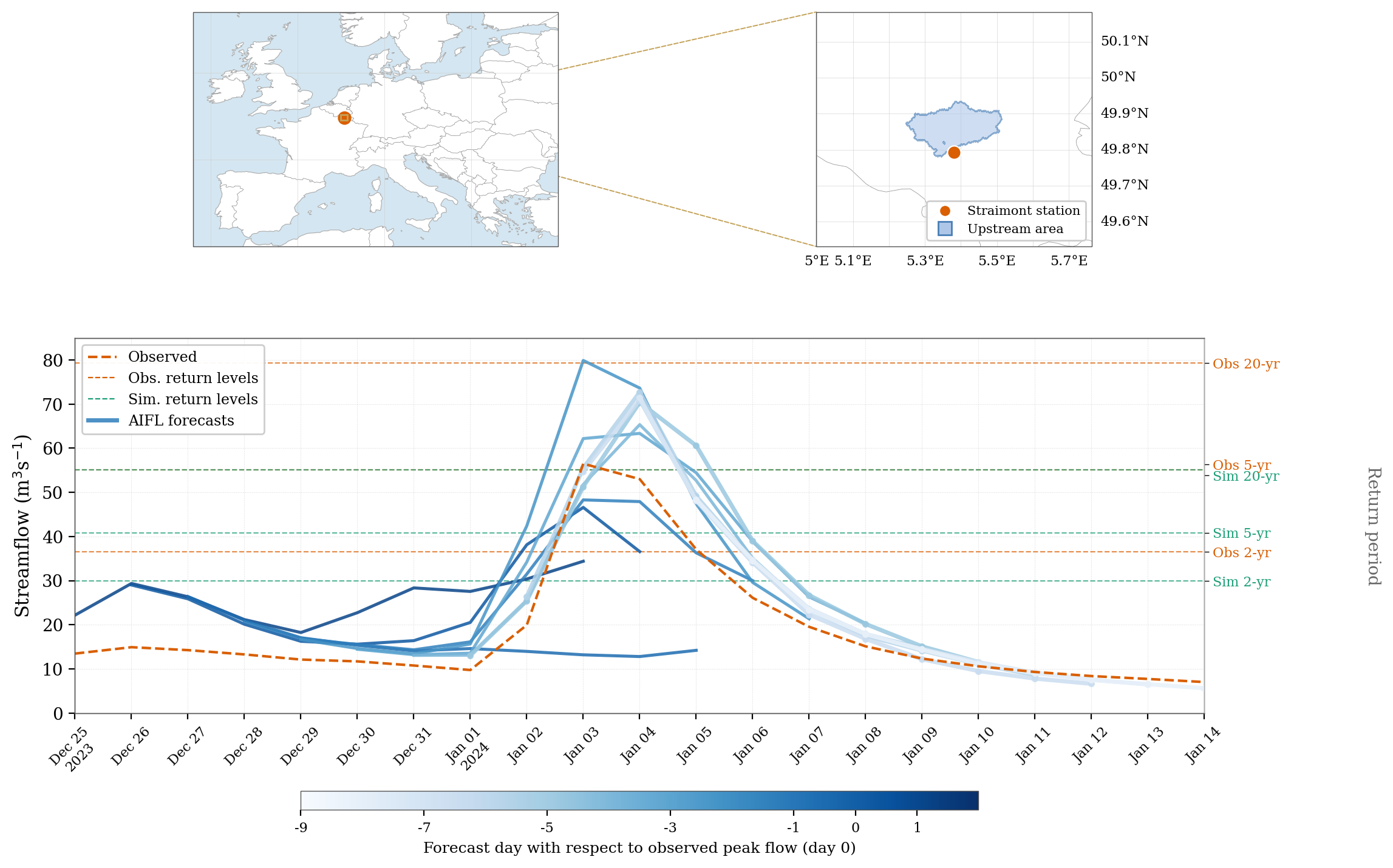}
\caption{\rev{Storm Henk case study (December 2023 -- January 2024) at the Straimont station, Belgium (GRDC ID 6221570; \SI{182}{\km\squared}). \textit{Top:} geographic context showing the station location and upstream catchment boundary. \textit{Bottom:} observed streamflow (orange dashed line) and ten-day AIFL forecasts issued on successive days (blue lines; colour indicates the forecast issue day relative to the observed peak on 3 January 2024, day~0). Horizontal dashed lines mark 2-, 5- and 20-year return-period thresholds derived from observed (orange) and simulated (green) streamflow climatologies; the systematic offset between the two sets reflects a negative bias in the model's long-term streamflow distribution.}}
\label{fig:case_study}
\end{figure}

While \revr{aggregate} statistics summarise global test-set performance, they do not fully convey operational realism during individual extreme events. To illustrate model behaviour in a real-world setting, performance is further examined through a focused case study of the January 2024 floods in Belgium associated with Storm Henk. Figure~\ref{fig:case_study} presents results for the Straimont station (GRDC ID 6221570), which drains a \SI{182}{\km\squared} catchment and experienced an approximately 1-in-5-year flood \rev{(based on observed return-period thresholds)}. \rev{It should be noted that this station was part of the ERA5-Land pre-training set (18,588 basins) but was not included in the IFS fine-tuning subset (2,010 basins). The model therefore operates in a partially out-of-distribution regime at this location.}

AIFL consistently predicts a clear flood signal six days in advance of the observed peak discharge. Early forecasts indicate peak magnitudes of approximately 50~m$^3$~s$^{-1}$ compared to an observed peak of approximately 57~m$^3$~s$^{-1}$, while subsequent forecast updates produce estimates in the range of 60--80~m$^3$~s$^{-1}$. \rev{The forecast initiated on December~27 notably fails to capture the subsequent flood peak\revr{. Analysis of the driving IFS precipitation forecast issued on that date shows that it initially predicted minimal rainfall over the catchment, followed by a sustained low precipitation rate before producing a distinct but delayed peak, leading to a substantial underestimation of the total accumulated precipitation that drives the flood signal}.} \rev{Conversely, close-range forecasts for January~3 (the observed peak day) are well calibrated, predicting approximately 51--56~m$^3$~s$^{-1}$ against the observed 56.5~m$^3$~s$^{-1}$; the 1-day-ahead forecast (55.5~m$^3$~s$^{-1}$) correctly exceeds the observed 5-year return-period threshold (55.1~m$^3$~s$^{-1}$), matching the observed exceedance.} \rev{However, close-range forecasts issued on January~1--3 overpredict discharge on January~4 by approximately 15--20~m$^3$~s$^{-1}$ (forecasting ${\sim}$70~m$^3$~s$^{-1}$ versus 53~m$^3$~s$^{-1}$ observed), indicating that the model delays the recession and overestimates a secondary peak that did not materialise.} Despite these uncertainties, the event is robustly detected well in advance, demonstrating the model's capacity for reliable early warning during high-impact floods. \rev{The figure also overlays observed and simulated return-period thresholds, which differ by up to ${\sim}$30\% at the 20-year level: the lower simulated thresholds would classify the observed peak as a $\geq$20-year event, whereas observation-based thresholds place it near the 5-year level, illustrating how threshold choice affects severity classification (see Section~\ref{subsec:flood_event_perf}).} \rev{This also affects false-alarm attribution for the January~4 overprediction ($\revr{\sim}$70~m$^3$~s$^{-1}$): \revr{the false alarm rate is dependent on both the return period and the threshold framework used.} At the 5-year level both frameworks flag a false alarm, since the forecast exceeds both the simulated and observed thresholds while the observation does not; at the 20-year level, only the dual framework registers a false alarm (forecast exceeds the simulated threshold but not the observed threshold.}

\revtwo{We note that this station exhibits a stronger negative bias than is typical of the test set: its simulated 20-year threshold lies about 30\% below the observed value, placing it among the most strongly underestimated basins (the lower ${\sim}$12\% of the distribution). Across the full test set the underestimation is considerably milder, with simulated thresholds averaging approximately 90\% of their observed counterparts at the 10- and 20-year return periods (see Section~\ref{subsec:flood_event_perf}). The Storm Henk case therefore represents a challenging, partially out-of-distribution example rather than the average behaviour of the model.}

\section*{Copyright}
© ECMWF 2026. This manuscript is published under a CC BY 4.0 license.

\section*{Code and Data Availability}
The CARAVAN and MultiMet datasets remain available via Zenodo and Google Cloud Storage as described in the original publications.

\section*{Author Contributions}
Following the CRediT taxonomy: \textbf{Maria Luisa Taccari}: Conceptualization, Methodology, Software, Formal analysis, Investigation, Writing -- original draft. 
\textbf{Kenza Tazi}: Case study design and analysis, Investigation, Writing -- review \& editing. 
\textbf{Oisín M. Morrison}: Data curation, Validation, Writing -- review. 
\textbf{Andreas Grafberger}: Data curation, Validation. 
\textbf{Juan Colonese}: Model operationalization, Software. 
\textbf{Corentin Carton de Wiart}: Model operationalization, Software. 
\textbf{Christel Prudhomme}: Project shaping, Supervision, Writing -- review. 
\textbf{Cinzia Mazzetti}: Project shaping, Writing -- review.
\textbf{Matthew Chantry}: Project shaping, Funding acquisition. 
\textbf{Florian Pappenberger}: Project shaping, Funding acquisition, Writing -- review.
All authors have read and agreed to the final version of the manuscript.

\section*{Acknowledgments}
The authors would like to thank Grey Nearing and Alden Keefe Sampson for the fruitful discussions that helped shape this work. The work presented in this paper has been produced in the context of the European Union’s Destination Earth Initiative and relates to tasks entrusted by the European Union to the European Centre for Medium-Range Weather Forecasts implementing part of this Initiative with funding by the European Union. Views and opinions expressed are those of the author(s) only and do not necessarily reflect those of the European Union or the European Commission. Neither the European Union nor the European Commission can be held responsible for them.

\section*{Declaration of generative AI and AI-assisted technologies in the manuscript preparation process}
During the preparation of this work the author(s) used Microsoft Copilot in order to rephrase sentences and improve the clarity of the text. After using these tools, the author(s) reviewed and edited the content as needed and take(s) full responsibility for the content of the published article.

\bibliographystyle{unsrt}
\bibliography{bibliography}

\end{document}